\documentclass[sigconf]{acmart}%

\usepackage{pdfpages}

\usepackage{booktabs} %

\usepackage{pifont}

\newcommand{\KS}[1]{\TODO{KS}{orange}{#1}}
\newcommand{\IC}[1]{\TODO{IC}{magenta}{#1}}
\newcommand{\DY}[1]{\TODO{DY}{orange}{#1}}
\newcommand{\MIN}[1]{\TODO{Min}{magenta}{#1}}
\newcommand{\J}[1]{\TODO{Julio}{orange}{#1}}
\newcommand{\D}[1]{\TODO{D}{red}{#1}}
\newcommand{\diego}[1]{\TODO{D}{red}{#1}}

\newcommand{\NEW}[1]{{\color{black}{#1}}} %
\newcommand{\NEWI}[1]{{\color{blue}{#1}}}%

\renewcommand{\KS}[1]{}
\renewcommand{\IC}[1]{}
\renewcommand{\DY}[1]{}
\renewcommand{\MIN}[1]{}
\renewcommand{\J}[1]{}
\renewcommand{\D}[1]{}
\renewcommand{\diego}[1]{}

\renewcommand{\NEW}[1]{{\color{black}{#1}}} %
\renewcommand{\NEWI}[1]{{\color{black}{#1}}}%

\graphicspath{{figs/}}

\usepackage[ruled,noend]{algorithm2e} %

\SetAlFnt{\small}
\SetAlCapFnt{\small}
\SetAlCapNameFnt{\small}
\SetAlCapHSkip{0pt}
\IncMargin{-\parindent}
\usepackage{ifpdf}
\usepackage{graphicx}
  \newcommand{\TODO}[3]{{\textcolor{#2}{\textbf{[\textsc{#1:} \textit{#3}]}}}}
\usepackage{color}
\definecolor{cyan}{cmyk}{1,0,0,0}
\definecolor{darkcyan}{rgb}{0.0,0.4,0.9}
\definecolor{darkgreen}{rgb}{0,0.5,0}
\definecolor{orange}{rgb}{1,0.5,0}
\definecolor{magenta}{cmyk}{0,1,0,0}
\definecolor{darkyellow}{cmyk}{0,0,0.75,0}
\definecolor{gray}{rgb}{0.8,0.8,0.8}

\usepackage[toc,page]{appendix} %
\usepackage{bm}
\usepackage{mathtools} %
\usepackage{nicefrac}

\usepackage{float}
\usepackage{soul}
\usepackage{array}
\usepackage{multirow}
\usepackage{setspace}
\usepackage{nicefrac}

\newcommand{\DELETE}[1]{} %
\newcommand{\IGNORE}[1]{}

\usepackage[ruled]{algorithm2e} %

\SetAlFnt{\small}
\SetAlCapFnt{\small}
\SetAlCapNameFnt{\small}
\SetAlCapHSkip{0pt}

\usepackage{microtype}
\acmJournal{TOG}

\copyrightyear{2023}
\acmYear{2023}
\setcopyright{rightsretained} 
\setcopyright{rightsretained}
\acmConference[SA Conference Papers '23]{SIGGRAPH Asia 2023 Conference
Papers}{December 12--15, 2023}{Sydney, NSW, Australia}
\acmBooktitle{SIGGRAPH Asia 2023 Conference Papers (SA Conference Papers
'23), December 12--15, 2023, Sydney, NSW, Australia}
\acmDOI{10.1145/3610548.3618140}
\acmISBN{979-8-4007-0315-7/23/12}

\newcommand{\xl}{\mathbf{x}_l}
\newcommand{\xs}{\mathbf{x}_s}

\newcommand{\xg}{\mathbf{x}_g}
\newcommand{\xv}{\mathbf{x}_v}

\def\norm#1{\left\|#1\right\|}
\def\d{\mathrm{d}}

\def\bfx{\mathbf x}

\def\sPath{\bar{\bfx}}

\def\sTime{\mathbf{t}}
\def\sTimesSpace{\mathcal{T}}

\def\tof{\mathrm{tof}}

\newcommand{\x}{\times}

\begin{document}

\title{Self-Calibrating, Fully Differentiable NLOS Inverse Rendering}

\author{Kiseok Choi}
\orcid{0000-0002-2352-3889}
\affiliation{%
  \institution{KAIST}
  \country{South Korea}}
\email{kschoi@vclab.kaist.ac.kr}

\author{Inchul Kim}
\orcid{0000-0003-1268-3104}
\affiliation{%
  \institution{KAIST}
  \country{South Korea}}
\email{ickim@vclab.kaist.ac.kr}

\author{Dongyoung Choi}
\orcid{0000-0003-1896-4038}
\affiliation{%
  \institution{KAIST}
  \country{South Korea}}
\email{dychoi@vclab.kaist.ac.kr}

\author{Julio Marco}
\orcid{0000-0001-9960-8945}
\affiliation{%
 \institution{Universidad de Zaragoza - I3A}
 \country{Spain}}
\email{juliom@unizar.es}

\author{Diego Gutierrez}
\orcid{0000-0002-7503-7022}
\affiliation{%
  \institution{Universidad de Zaragoza - I3A}
  \country{Spain}}
\email{diegog@unizar.es}

\author{Min H. Kim}
\orcid{0000-0002-5078-4005}
\affiliation{%
  \institution{KAIST}
  \country{South Korea}}
\email{minhkim@vclab.kaist.ac.kr}

\begin{abstract}
Existing time-resolved non-line-of-sight (NLOS) imaging methods reconstruct hidden scenes by inverting the optical paths of indirect illumination measured at visible relay surfaces.
These methods are prone to reconstruction artifacts due to inversion ambiguities and capture noise, which are typically mitigated through the manual selection of filtering functions and parameters. 
We introduce a fully-differentiable end-to-end NLOS inverse rendering pipeline that self-calibrates the imaging parameters during the reconstruction of hidden scenes, using as input only the measured illumination while working both in the time and frequency domains.
Our pipeline extracts a geometric representation of the hidden scene from NLOS volumetric intensities and estimates the time-resolved illumination at the relay wall produced by such geometric information using differentiable transient rendering. 
We then use gradient descent to optimize imaging parameters by minimizing the error between our simulated time-resolved illumination and the measured illumination. 
\NEWI{Our end-to-end differentiable pipeline couples diffraction-based volumetric NLOS reconstruction with path-space light transport and a simple ray marching technique to extract detailed, dense sets of surface points and normals of hidden scenes.}
\NEWI{We demonstrate the robustness of our method to consistently reconstruct geometry and albedo, even under significant noise levels.} 
\end{abstract}



\maketitle

\section{Introduction}
\label{sec:intro}

Time-gated non-line-of-sight (NLOS) imaging algorithms aim to reconstruct hidden scenes by analyzing time-resolved indirect illumination on a visible relay surface \cite{jarabo2017recent,Satat2016,Faccio2020non}. These methods typically emit ultra-short illumination pulses on the relay surface, and estimate the hidden scene based on the time of flight of third-bounce illumination reaching the sensor \cite{Velten2012nc,OToole:2018:confocal,Lindell2019wave,Xin2019theory,Liu:2019:phasor}. 

\begin{figure}
	\centering
	\includegraphics[width=0.89\linewidth]{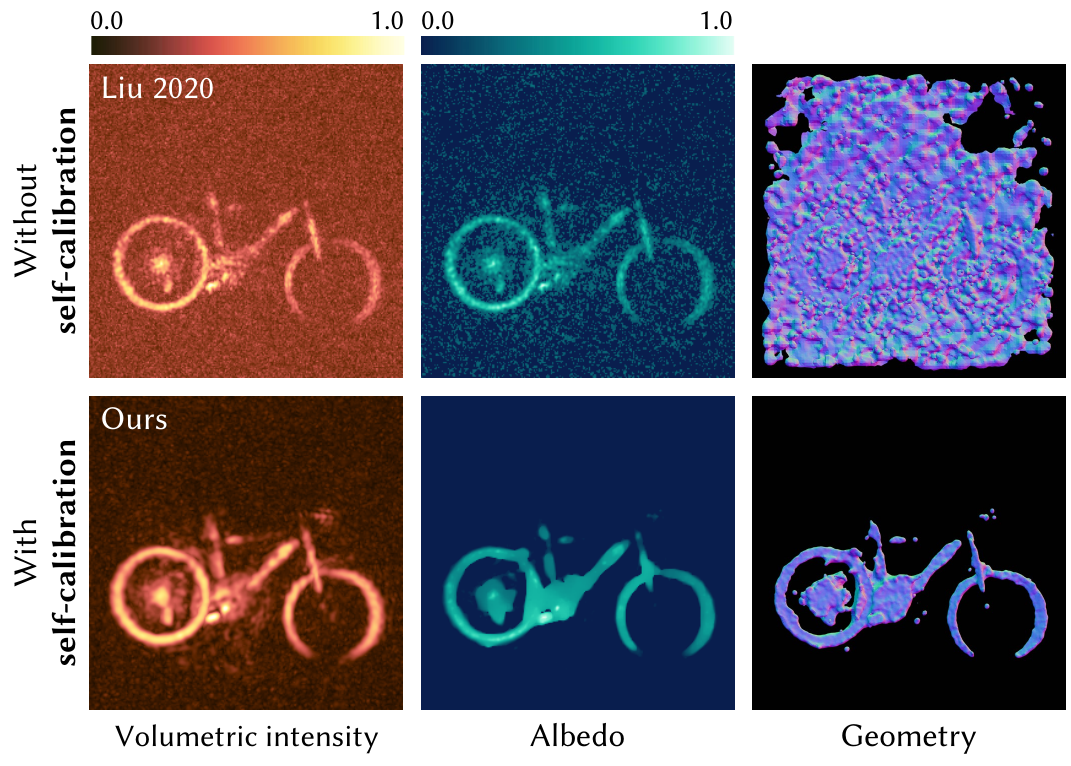}%
	\caption[]{\label{fig:teaser}%
We present a self-calibrating, fully-differentiable NLOS inverse rendering pipeline for the reconstruction of hidden scenes. Our method only requires transient measurements as input and relies on differentiable rendering and implicit surface estimation from NLOS volumetric outputs to obtain the optimal NLOS imaging parameters that yield accurate surface points, normals, and albedo reconstructions of the hidden scene. 
The top row shows the reconstructed volumetric intensity, albedo, and 3D geometry of a real scene \cite{Liu:2020:phasor}, failing to reconstruct geometry estimation due to noise interference. The bottom row demonstrates our results after optimization of the imaging parameters.}
\end{figure}

\NEW{The majority of existing methods estimate hidden geometry by backprojecting captured third-bounce illumination into a voxelized space that represents the hidden scene \cite{laurenzis2014feature}, lacking information about surface orientation and self-occlusions \cite{iseringhausen2020non}. Moreover, captured data contains higher-order indirect illumination and high-frequency noise from different sources that introduce undesired artifacts in the reconstructions. 
Performing a filtering step over the data or the reconstructed volume is the most common solution to mitigate errors and enhance the geometric features \cite{Arellano2017NLOS,Velten2012nc, Buttafava2015, OToole:2018:confocal, Liu:2019:phasor}; however, this requires manual design and selection of filter parameters, as their impact in the reconstruction quality is highly dependent on the scene complexity, environment conditions, and hardware limitations.}

Recent physically-based methods proposed an alternative technique that avoids the issues linked to backprojection. 
By merging a simplified but efficient three-bounce transient rendering formula with an optimization loop, 
the computed time-resolved illumination at the relay wall \NEW{resulting from an optimized geometry reconstruction} is compared to the measured illumination. %
\NEWI{However, geometric representations introduced by existing works limit the detail in the reconstructions \cite{iseringhausen2020non} or fail to reproduce the boundaries of hidden objects \cite{tsai2019beyond}.}

Alternatively, the recent development of accurate transient rendering methods \cite{Jarabo2014,pediredla2019ellipsoidal,royo2022non} has fostered differentiable rendering pipelines in path space \cite{wu2021differentiable,Yi2021SIGA}, which have the potential to become key tools in optimization schemes. However, differentiable methods are currently bounded by memory limitations since the need to compute the derivatives of time-resolved radiometric data severely limits the number of unknown parameters that can be handled. \NEW{The difficulty of handling visibility changes in a differentiable manner, as well as the large number of parameters that need to be taken into account, are two limiting factors shared as well with steady-state differentiable rendering \cite{li2018differentiable,zhao2020physics}, that are further aggravated in the transient regime.  As a result, NLOS imaging methods that rely on differentiable rendering are therefore limited to simple operations such as tracking the motion of a single hidden object with a known shape \cite{Yi2021SIGA}.}

To address these problems, we propose a novel self-calibrated, fully differentiable pipeline for NLOS inverse rendering that jointly optimizes system parameters and scene information to extract surface points, normals, and albedo of the hidden geometry. 
To this end, we combine diffractive phasor-field imaging in the frequency domain \cite{Liu:2019:phasor,Liu:2020:phasor} with differentiable third-bounce transient rendering in the temporal domain. 
\NEWI{We leverage the volumetric output of phasor-field NLOS imaging to estimate geometric information of the hidden scene, which we then use on a transient rendering step to simulate time-resolved illumination at the relay wall. By minimizing the error between simulated and captured illumination, we provide a fully-differentiable pipeline for self-calibrating NLOS imaging parameters in an end-to-end manner.}

\NEW{Our optimized parameters provide accurate volumetric outputs from which we estimate surface points, normals and albedos of hidden objects, with more geometric detail than previous surface-based methods. Our method is robust in the presence of noise, providing consistent geometric estimations under varying capture conditions.}
\NEWI{Our code is freely available for research purposes\footnote{\url{https://github.com/KAIST-VCLAB/nlos-inverse-rendering.git}}.}

\section{Related Work}
\label{sec:relatedwork}
Active-light NLOS imaging methods provide 3D reconstructions of general NLOS scenes by leveraging temporal information of light propagation by means of time-gated illumination and sensors \cite{Faccio2020non,jarabo2017recent}. 

\paragraph{Scene representation}
\NEW{While existing methods rely on inverting third-bounce transport, they may differ in their particular representation of scene geometry as volumetric density or surfaces. 
Volumetric approaches estimate geometric density by backprojecting third-bounce light paths onto a voxelized space \cite{Velten2012nc, Gupta2012, Buttafava2015, Gariepy2015, Arellano2017NLOS, ahn2019convolutional, LaManna2018error}.}
Efficiently inverting the resulting discrete light transport matrix is not trivial; many dimensionality reduction methods have been proposed \cite{Lindell2019wave, Xin2019theory, OToole:2018:confocal, Young2020Directional, heide2019non}, but they are often limited in spatial resolution (as low as 64$\x$64 in some cases) due to memory constraints. 
Surface methods, in contrast, rely on inverting third-bounce light transport onto explicit representations of the geometry \cite{tsai2019beyond,iseringhausen2020non,plack2023fast}, usually starting with simple blob shapes, 
\NEWI{progressively optimizing the geometry until loss converges.}
\NEWI{In contrast, we estimate \textit{implicit} geometric representations of the hidden scene based on surface points and normals by ray marching the volumetric output of NLOS imaging, inspired by recent work on neural rendering \cite{mildenhall2020nerf, barron2021mipnerf, Niemeyer2021Regnerf}.}
\NEWI{The combination of NLOS imaging with differentiable transient rendering over the estimated geometry allows us to self-calibrate imaging parameters in an end-to-end manner.} 
\NEWI{For clarity, in this paper the term \emph{explicit} surface refers to a polygonal surface mesh, while \emph{implicit} surface denotes a representation based on surface points and their normals, without defining a surface mesh.} Please, refer to Section~\ref{sec:implicit-geometry} for a further detailed discussion on explicit/implicit surface representations.

\begin{figure}[t]
	\centering
	\includegraphics[width=1.0\linewidth]{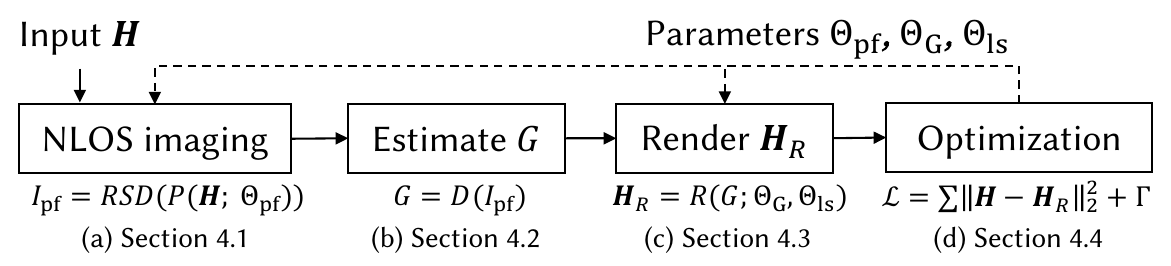}\\%
	\caption[]{\label{fig:full_pipeline}%
		Overview of our self-calibrated, fully differentiable NLOS inverse rendering workflow (Sections \ref{sec:NLOS-imaging} and \ref{sec:ourmethod}).
		(a) We perform NLOS imaging using a phasor-field diffraction method, taking an initial matrix ~$\bm{H}$ of transient measurements as input, and outputting volumetric intensity $I_\mathrm{pf}$.
		(b) We estimate ~$G$, an implicit geometric representation of the hidden scene, from $I_\mathrm{pf}$.
		(c) We obtain the time-resolved illumination $\bm{H_R}$ from $G$ using differentiable path-space transient rendering.
		(d) We optimize imaging parameters until the error between $\bm{H}$ and $\bm{H_R}$ converges \NEW{with regularization terms~$\Gamma$.}
		Geometry {$G$} is computed during the forward pass, while {$\Theta_\mathrm{pf}$}, {$\Theta_\mathrm{ls}$}, and {$\Theta_\mathrm{G}$} are updated during the backward pass.}
\end{figure}

\paragraph{Learning-based approaches}
Other methods leverage neural networks instead, such as U-net \cite{GrauCVPR2020}, convolutional neural networks \cite{Chen2020learned}, or neural radiance fields \cite{Mu:ICCP:2022:physics}.
These learning-based methods are learned using object databases such as 
ShapeNet~\cite{chang2015shapenet}.
However, their parameters are trained with steady-state renderings of synthetic scenes composed of a single object behind an occluder in an otherwise empty space. As such, their performance is often degraded with real scenes, often overfitting to the training dataset, and becoming susceptible to noise.  
Our method does not rely on a pre-trained deep network to extract high-level features from synthetic steady-state rendering data; instead, we explicitly optimize virtual illumination functions and scene information by evaluating actual transient observations, without relying on neural networks. 
\NEWI{Recent works by \citet{shen2021non} and \citet{Fujimura2023} leverage transient observations similar to ours for optimizing multi-layer perceptrons for imaging. 
However, these methods cannot be utilized for calibrating the filtering parameters of volumetric NLOS methods due to the lack of evaluation of the physical observation of the transient measurements by an NLOS imaging and light transport model.}

\paragraph{Wave-based NLOS imaging}
Recent works have shifted the paradigm of third-bounce reconstruction approaches to the domain of wave optics \cite{Liu:2019:phasor, Lindell2019wave}. 
\NEW{In particular, the phasor field framework \cite{Liu:2019:phasor} computationally transforms the data captured on the relay surface into illumination arriving at a virtual imaging aperture}. This has enabled more complex imaging models (e.g., \cite{Marco2021NLOSvLTM,doveNonparaxialPhasorfieldPropagation2020,doveSpeckledSpeckledSpeckle2020,Guillen2020Effect,rezaPhasorFieldWaves2019a}\NEWI{)}, 
and boosted the efficiency of NLOS imaging to interactive and real-time reconstruction rates \cite{Liu:2020:phasor,nam2021low,liao2021fpga,Mu:ICCP:2022:physics}. However, these systems require careful calibration of all \NEWI{their} parameters, including the definition of the phasor field and the particular characteristics of lasers and sensors, which makes using them a cumbersome process. Our fully self-calibrated system overcomes this limitation.

\section{Time-gated NLOS imaging model}
\label{sec:NLOS-imaging}
We propose a differentiable end-to-end inverse rendering pipeline (shown in Figure \ref{fig:full_pipeline}) to improve the reconstruction quality of hidden scenes by optimizing the parameters of NLOS imaging algorithms without prior knowledge of the hidden scene. In the following, we describe our NLOS imaging model. Section \ref{sec:ourmethod} describes our optimization pipeline based on this NLOS imaging model.

\subsection{Phasor-based NLOS imaging}
\label{sec:phasor-fields}
\begin{figure}[tp]
	\centering
	\vspace{-2.5mm}%
	\includegraphics[width=0.9\linewidth]{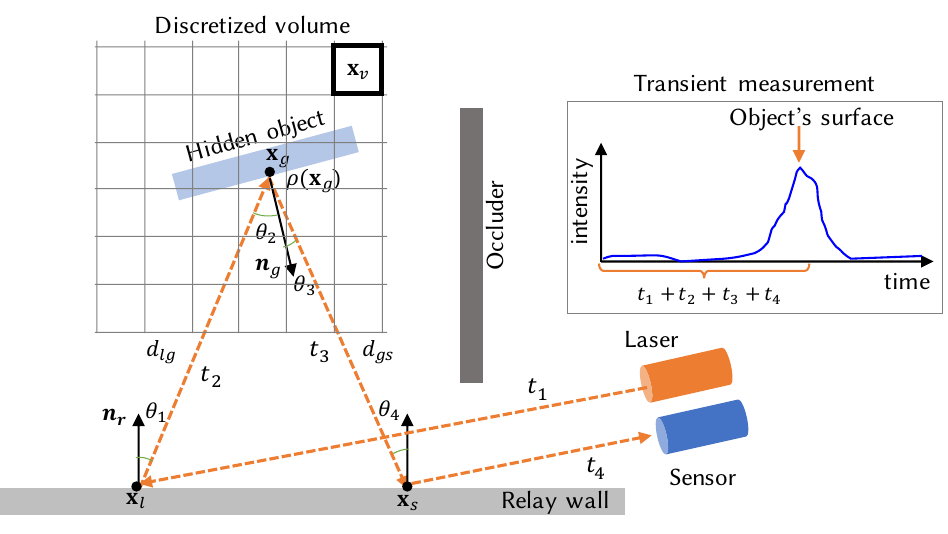}%
	\caption[]{\label{fig:nlos_setup}
		NLOS imaging setup.
		A laser emits a pulse of light, which travels to the relay wall, then to the hidden geometry, back to the relay wall, and reaches the sensor after a travel time of ~$t = t_1 + t_2 + t_3 + t_4$.
		The inset shows the sensor response; the peak at~$t$ indicates the presence of a hidden object.
	}	
\end{figure}

In a standard NLOS imaging setup (see Figure~\ref{fig:nlos_setup}), a laser beam is emitted towards a point $\xl$ on a visible relay wall, which reflects light towards the hidden scene and then is reflected back to the wall. The hidden scene is imaged based on the time of flight of the time-resolved illumination, captured at points $\xs$ on the relay wall in the form of a measurement matrix $\bm{H}(\xl,\xs,t)$.

The recent diffractive phasor-based framework by Liu et al. \shortcite{Liu:2019:phasor,Liu:2020:phasor} intuitively turns the grid of measured points $\xs$ on the relay wall into a virtual aperture; this allows to formulate the reconstruction of NLOS scenes as a virtual \textit{line-of-sight} (LOS) problem.

We define $\bm{H}\left(\mathbf{x}_l, \mathbf{x}_s,\Omega \right)$ as a set of phasors at the relay wall, 
obtained by Fourier transform of the measurement matrix $\bm{H}(\xl,\xs,t)$. 
In practice, since this function $\bm{H}$ is noisy, we apply a filtering operation as 
\begin{align}
	\label{eq:rsd_illumination}
	\bm{H}_\mathrm{pf}\left(\mathbf{x}_l, \mathbf{x}_s,\Omega \right) = \mathcal{P}\left(\mathbf{x}_l,\mathbf{x}_s,\Omega \right) \bm{H}\left(\mathbf{x}_l, \mathbf{x}_s,\Omega \right),
\end{align}
where $\mathcal{P}(\xl,\xs,\Omega)$ represents a virtual illumination function that acts as a filter over $\bm{H}$, typically defined as a spatially-invariant illumination function \cite{Liu:2019:phasor,Liu:2020:phasor}. The hidden scene can then be imaged as an  intensity function $I_\mathrm{pf}(\xv,t)$ on a voxelized space via Rayleigh-Sommerfeld Diffraction (RSD) operators as 
\begin{align}
	\label{eq:rsd}
 I_\mathrm{pf}\left( \mathbf{x}_v,t \right) = 
	\resizebox{0.87\linewidth}{!}{
		\mbox{\fontsize{10}{12}\selectfont $
			\left | \int\limits^{\infty}_{-\infty} e^{i\frac{\Omega}{c}{t}} \int\limits_S \!\! \int\limits_L \frac{e^{-i\frac{\Omega }{c} (d_{lv}+d_{vs})}}{ d_{lv} d_{vs} } \bm{H}_\mathrm{pf}\left(\mathbf{x}_l, \mathbf{x}_s,\Omega \right) \d \mathbf{x}_l \d \mathbf{x}_s \frac{\d\Omega}{2\pi} \right |^2,
			$ } } %
\end{align}
where 
$L$ and $S$ define the illuminated and measured regions on the relay wall, respectively;
$d_{lv} = \left\| \xl - \xv \right\|$ and $d_{vs} = \left\| \xv - \xs \right\|$ are voxel-laser and voxel-sensor distances (see Figure~\ref{fig:nlos_setup}); and $\Omega$~represents frequency.

Classic NLOS reconstruction methods 
reconstruct hidden geometry by evaluating $\bm{H}(\xl,\xs,t)$ at the time of flight of third-bounce illumination paths between scene locations and points on the relay surface \cite{OToole:2018:confocal,Arellano2017NLOS,Gupta2012}. This is analogous to evaluating $I_{\mathrm{pf}}(\xv,t)$ at $t=0$, where the RSD propagators have traversed an optical distance $\|\sPath\| = d_{lv}+d_{vs}$. 
We incorporate a similar third-bounce strategy in our path integral formulation as described in the following. Due to the challenges of estimating surface albedo due to diffraction effects during the NLOS imaging process \cite{Guillen2020Effect,Marco2021NLOSvLTM}, we assume an albedo term per surface point that approximates the averaged reflectance observed from all sensor points.

\subsection{Path-space light transport in NLOS scenes}
\label{sec:NLOS-transport}

To formally describe transient light transport in an efficient manner, we rely on the transient path integral formulation \cite{Jarabo2014,royo2022non}. 
Transient light transport 
\NEWI{$\bm{H}(\xl,\xs,t) \in \mathbb{R}$} can then be expressed as 
\begin{align}
\label{eq:transient-irradiance}
\bm{H}(\xl,\xs,t)= \int_\sTimesSpace \int_\psi \mathcal{K}(\sPath,\sTime) \d\mu(\sPath) \d\mu(\sTime),
\end{align}%
where $\mathcal{K}$ is the radiometric contribution in transient path-space;
$\d\mu(\sPath)$ is the differential measure of path $\sPath$;
$\sTimesSpace$ represents the domain of temporal measurements; 
$\sTime=t_l\dots t_s$ is the sequence of time-resolved measurements on each vertex;
$\d\mu(\sTime)$ denotes temporal integration at each vertex;
$\sPath=\xl \dots \xs$ is a set of discrete transient path time intervals of $k+1$ vertices;
and $\psi = \cup_{k=1}^\infty \psi_k$ is the entire space of paths with any number of vertices, with $\psi_k$ being the space of all paths with $k$ vertices. For convenience and without losing generality, we ignore the fixed vertices at the laser and sensor device in our formulae. 

In practice, $\bm{H}$ is obtained by the spatio-temporal integration of transient measurements during a time interval $\tau$, %
which accounts for the contribution of all paths $\sPath$ with time of flight
\begin{align}
	\label{eq:tof-function}
	t = \tof(\sPath)=\sum\nolimits_{i=1}^{k}\frac{||\bfx_i-\bfx_{i-1}||}{c},
\end{align}
where $c$ is the speed of light, $\bfx_0 \equiv \xl$, and $\bfx_k \equiv \xs$. We assume no scattering delays at the vertices.

\begin{figure*}[t]
	\centering
	\includegraphics[width=0.95\linewidth]{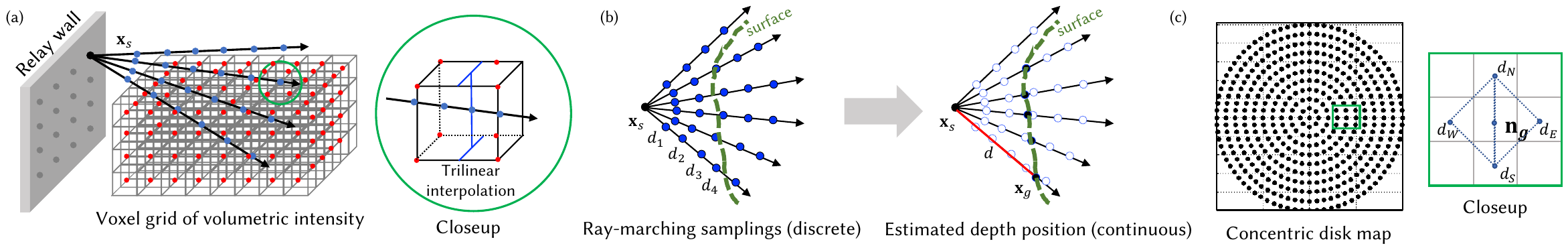}%
	\caption[]{\label{fig:geom_all}
		{Geometry estimation procedure. (a) We ray-march from sensor points $\bfx_s$, and estimate the intensity at each point along the ray by trilinear interpolation of $I_\mathrm{pf}$. (b) From the discrete ray-marching samplings, we obtain a continuous depth function. (c) Normals are computed based on the distances at neighboring ray samples in the concentric hemispherical mapping. }
	}
	\label{fig:full_geometry_reconstruction}
\end{figure*}

Incorporating the third-bounce strategy of NLOS reconstruction methods in our path integral formulation, we can express $\mathcal{K}$ in a closed form as %
\begin{align}
\label{eq:transient-radiance}
	\mathcal{K}(\sPath,\sTime) = \Lambda (\xl \to \xg, t_l)  \rho(\xg) \mathfrak{T}(\sPath,\sTime) \Phi (\xg\to\bfx_s, \tof(\sPath)) ,
\end{align}
where $\Lambda$ is the emitted light from the laser, $\Phi$ is the time-dependent sensor sensitivity function,  $\rho$ represents surface reflectance, and $\mathfrak{T}(\sPath,\sTime)$ is the path throughput defined by
\begin{equation}
	\label{eq:tran_path_throughput}
	\mathfrak{T}(\sPath,\sTime) = V(\xl, \xg)\frac{|{\cos {\theta _1}||\cos {\theta _2}}|}{{d_{lg}^2}}V(\xg, \xs)\frac{|{\cos {\theta _3}||\cos {\theta _4}}|}{{d_{gs}^2}},
\end{equation}
where $V$ is the binary visibility function between two vertices, 
$d_{lg} = \|\mathbf{x}_l-\mathbf{x}_g\|$ and $d_{gs} = \|\mathbf{x}_g-\mathbf{x}_s\|$,
and $\theta_{1-4}$ refer to the angles between the normals of both the relay wall and surface geometry, and the path segments in $\sPath$ (see Figure~\ref{fig:nlos_setup}). Note that the three-bounce illumination is expressed in the path space as $\sPath \equiv \xl \rightarrow \xg \rightarrow \xs$.

	\NEW{Neither the emitted light $\Lambda$ nor the sensor sensitivity $\Phi$ are ideal Dirac delta functions. \citet{Yi2021SIGA} and  \citet{Hernandez2017SPAD} provide the following models for the laser and sensor behavior}
	\begin{align}
		\label{eq:laser-model2}
		\NEW{{\Lambda}(t)} & \NEW{= \frac{I_l}{{\sigma_l \sqrt {2\pi } }} {e^{ - {{ {t } }^2}/( {2{\sigma_l ^2}} )}}},\\ 
		\label{eq:sensor-model2}
		\NEW{\Phi(t)} &  \NEW{={\kappa_s {e^{ - \kappa_s t}}} * \frac{1}{{\sigma_s \sqrt {2\pi } }} {e^{ - {{ {(t-\mu_s)} }^2}/( {2{\sigma_s ^2}} )}}},
	\end{align}
	\NEW{where $\sigma_l$ is the standard deviation of the Gaussian laser pulse, $I_l$~is the laser intensity, $\kappa_s$ is the sensor sensitivity decay rate, 
	$\sigma_s$ is the standard deviation of the sensor jitter, and $\mu_s$ is the offset of the sensor jitter. %
	Since we are only interested on reproducing the combined effect of the laser and sensor models $\Lambda$ and $\Phi$ on the path throughput (Equation~\ref{eq:tran_path_throughput}), we replace them by a single joint laser-sensor correction function as}
\begin{align}
\begin{aligned}
\label{eq:joint-laser-sensor-model}
\NEW{\Psi(t)} & \NEW{= \Phi(t) * \Lambda(t)} \\
		& \NEW{= {\kappa_s {e^{ - \kappa_s t}}} * \frac{I_l}{{\sigma_{ls} \sqrt {2\pi } }} {e^{ - {{ t }^2}/( {2{\sigma_{ls} ^2}} )}}.}
\end{aligned}
\end{align}
\NEW{Note that the convolution of the two Gaussian functions of Equations~\ref{eq:laser-model2} and~\ref{eq:sensor-model2} yields a single Gaussian with a joint model parameter $\sigma_{ls}=\sqrt{{\sigma}_{l}^2+{\sigma}_{s}^2}$. 
We set the sensor jitter offset as $\mu_s = 0$, with the assumption that a uniform distribution of shifts is equally present in all transient measurements.
Please refer to the supplemental material for more details on derivation.
Our inverse rendering optimization seeks optimal parameters of this model automatically based on physically-based transient rendering.}

\section{Differentiable Time-gated NLOS Inverse Rendering}
\label{sec:ourmethod}
\NEW{In the following, we describe in detail our self-calibrated, end-to-end differentiable inverse rendering pipeline}, where the forward pass provides high-detailed reconstructions of the geometry $G$, while the backward pass optimizes per-voxel surface reflectance as albedo~$\Theta_G$, as well as system parameters $\Theta_\mathrm{pf}$ and $\Theta_\mathrm{ls}$ to improve the forward pass reconstruction. \NEW{For clarity, from here on, we redefine our functions in terms of their parameters to be optimized.} \NEW{Refer to the supplemental material for a summary of the different symbols.}

\subsection{Virtual illumination for RSD propagation}
\label{sec:virtual-illumination}

The inputs to our system are the known locations of the illumination~$\xl$ and the sensor $\xs$, a matrix ~$\bm{H}$ of transient measurements, and an \textit{arbitrary} virtual illumination function~\NEW{$\mathcal{P}(\Theta_\mathrm{pf}) \equiv \mathcal{P}(\xl,\xs,\Omega)$ (Equation~\ref{eq:rsd_illumination}), where $\Theta_\mathrm{pf}$ represents the optimized parameter space for $\mathcal{P}$}. 
Based on previous works \cite{Liu:2019:phasor,Liu:2020:phasor,Marco2021NLOSvLTM}, we define $\Theta_\mathrm{pf} = \{\sigma_\mathrm{pf}, \Omega_\mathrm{pf} \}$ to model a central frequency with a zero-mean Gaussian envelope as 
$\mathcal{P}(\Theta_\mathrm{pf})= e^{i\Omega_\mathrm{pf} t} e^{{-t^2} / {(2\sigma_\mathrm{pf}^2)}}$,
where $\sigma_\mathrm{pf}, \Omega_\mathrm{pf}$ represent the standard deviation and central frequency, respectively.
Note that this equation is fully differentiable. 
\NEW{In the forward pass we first compute the filtered matrix $\bm{H}_\mathrm{pf}$ (Equation~\ref{eq:rsd_illumination}) 
using the optimized virtual illumination $\mathcal{P}(\Theta_\mathrm{pf})$, having $\bm{H}_\mathrm{pf} = P(\bm{H}; \Theta_\mathrm{pf})$ (Figure~\ref{fig:full_pipeline}a).}
We then compute a first estimation of the volumetric intensity $I_\mathrm{pf}$ of the hidden scene by evaluating RSD propagation (Equation~\ref{eq:rsd}) at $t=0$, as $I_\mathrm{pf} = RSD(\bm{H}_\mathrm{pf})$.
Next, we show how to estimate both the geometry $G$ and the time-resolved transport $\bm{H}_R$ at the relay wall.

\subsection{\NEW{Implicit surface geometry}}
\label{sec:implicit-geometry}

Our next goal is to estimate an implicit surface representation~$G$ (points $\xg$ and normals $\mathbf{n}_g$) by means of a differentiable function $D$ as $G = D(I_\mathrm{pf})$ \NEW{(Figure \ref{fig:full_pipeline}b) that takes our volumetric intensity function~$I_\mathrm{pf}$ as input}.

We keep an implicit representation of our hidden surface geometry~$G$ without creating meshed (explicit) surface geometry during the whole optimization. %
The key idea is to use the volumetric data computed at each forward pass to estimate \emph{projections} of the geometry (i.e., points and normals) visible from the perspective of each sensor point~$\xs$ on the relay wall and use those to perform path-space differentiable transient rendering at~$\xs$. 

We first estimate the geometry observed by $\xs$ by sampling rays towards our volumetric intensity~$I_\mathrm{pf}$, and build an implicit representation of the closest surface along each ray.
Using information from neighboring rays, we then estimate the normals required to compute the path-space throughput of $\mathfrak{T}$ (Equation~\ref{eq:tran_path_throughput}). 
Using the implicit geometry computed for every sensing point $\xs$, we then compute time-resolved illumination at~$\xs$ as we describe later in this subsection.

\paragraph{Points}
As Figure~\ref{fig:full_geometry_reconstruction}a shows, for each sensor point $\xs$ we sample rays uniformly using concentric hemispherical mapping \cite{shirley1997low}. 
We then sample points along each ray with ray marching, and estimate the intensity at each sampled point (blue in Figure~\ref{fig:full_geometry_reconstruction}a) by trilinear interpolation of neighbor voxel intensities of~$I_\mathrm{pf}$ (red).
From the interpolated volumetric intensities $I_\mathrm{pf}(d_i)$ (Figure~\ref{fig:full_geometry_reconstruction}b, left), we estimate the distance $d_{gs}$ between $\xs$ and the hidden surface vertex $\xg$ (Figure~\ref{fig:full_geometry_reconstruction}b, right), assuming $\xg$ is located at the maximum intensity along the ray.
To find $d_{gs}$ in free space from the ray-marched intensities in a differentiable manner, we use {\tt{softargmax}} function: 
$d_{gs} = \frac{{\sum\nolimits_i { \omega_i {d_i} } }}{{\sum\nolimits_i  \omega_i}}$,
where $d_i$ is a ray-marched distance from $\xs$, and $\omega_i = {e^{\beta {I_{\mathrm{pf},i}}}}$ is a \NEWI{probability} density function of $d_i$, 
and $I_{\mathrm{pf},i}$ is the volume intensity at distance $d_i$ along the ray. 
$\beta$ is a hyperparameter that determines the sensitivity in blending neighboring probabilities, set to 1e+3 in all our experiments. 
If $I_{\mathrm{pf}}$ falls below a threshold, we assume that no surface has been found; we set this threshold to 
0.05 for synthetic scenes, and 0.2 for real scenes throughout the paper.
Our procedure implicitly estimates surface points~$\xg$ at distances $d=\norm{\xs-\xg}$ by observing via ray marching the grid of phasor-field intensities $I_{\mathrm{pf}}$ from the perspective of the sensing points $\xs$.

\paragraph{Normals}
As shown in Figure~\ref{fig:full_geometry_reconstruction}c, we estimate the normal $\mathbf{n}_g$ at vertex $\xg$  based on the distances $d_N, d_S, d_E, d_W$ at neighboring ray samples in the concentric hemispherical mapping. We compute the normals of two triangles $\triangle d_N d_E d_S$ and $\triangle d_S d_W d_N$ via two edges' cross product and compute $\mathbf{n}_g$ as the normalized sum of the normals of those two triangles.

\paragraph{Surface albedo}
Besides points and normals---updated implicitly during each forward pass---, computing path contribution $\mathcal{K}$ (Equation~\ref{eq:transient-radiance}) at sensor points $\xs$ requires computing per-point \NEW{monochromatic albedo} $\rho$. 
\NEWI{We estimate albedos by evaluating the physical observation of
 the transient measurements in the backward pass.}

\begin{figure*}[t]
	\centering
	\includegraphics[width=0.85\linewidth]{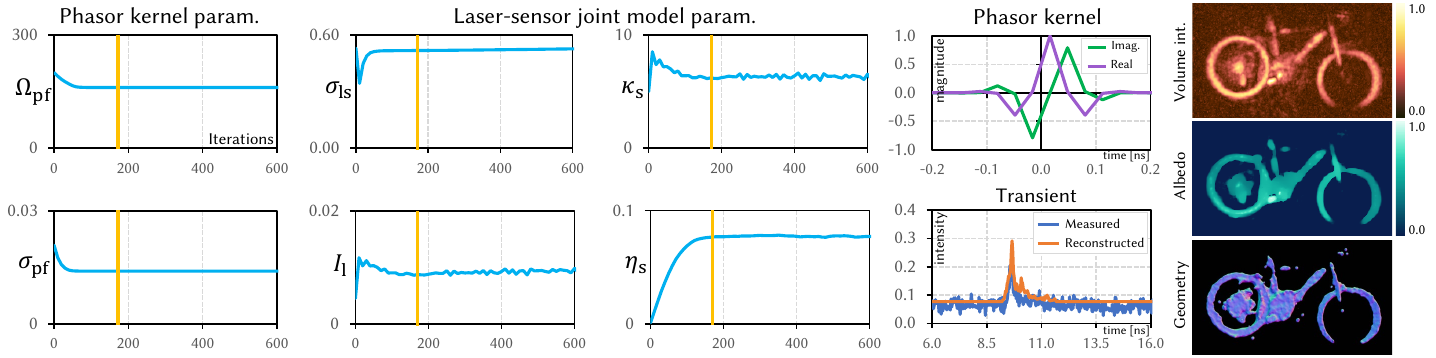}%
	\caption[]{\label{fig:converge-plot}
		\NEW{Convergence of the imaging parameters optimized by our method in the \textsc{Bike} real scene. From left to right: Phasor kernel parameters ($\Omega_\mathrm{pf}$, $\sigma_\mathrm{pf}$), laser-sensor joint model parameters ($\sigma_\text{ls}$, $I_\text{l}$, $\kappa_\text{s}$, $\eta_\text{s}$),} the converged phasor kernel (purple and green for real and imaginary parts), measured transients compared to our reconstructed one, and our reconstruction results after the optimization. The yellow line indicates when the optimization converges. The converged phasor kernel yields a high-quality reconstruction, while the laser and sensor parameters provide an accurate estimation of transient illumination.}
\end{figure*}

\subsection{Differentiable transient rendering}
\label{sec:transient_transport}
The next step during the forward pass is to obtain time-resolved illumination $\bm{H}_R$ at $\xs$ through transient rendering. 
In our pipeline (Figure~\ref{fig:full_pipeline}c), we represent this step as $\bm{H}_R = R(G;\Theta_\mathrm{G}, \Theta_\mathrm{ls})$,
where $R()$ computes third-bounce time-resolved light transport at sensing points $\xs$.
We use the rays sampled from $\xs$ (Figure~\ref{fig:full_geometry_reconstruction}b) to compute the radiometric contribution $\mathcal{K}(\sPath,\sTime)$ of the implicit surface points $\xg$ estimated by those rays, following Equations~\ref{eq:transient-radiance} through \ref{eq:joint-laser-sensor-model}. 

\paragraph{Visibility}
\NEW{Differentiating the binary visibility function $V$, necessary to compute the path throughput $\mathfrak{T}$ (Equation~\ref{eq:tran_path_throughput}), is challenging.} 
\NEWI{However, note that we estimate an implicit surface at $\xg$ based on volumetric intensities, which strongly depend on the illumination from the laser reaching the surface and going back to the sensor without finding any occluder. Based on this, we avoid computing the visibility term by assuming the volumetric intensities are a good estimator of the geometry visible from the perspective of both laser and sensor positions on the relay wall.}

\paragraph{Transient rendering} 
The radiometric contribution $\mathcal{K}(\sPath,\sTime)$ (Equation \ref{eq:transient-radiance}) yields time-resolved transport in path space for a single path  $\sPath \equiv \xl \rightarrow \xg \rightarrow \xs$. Our goal is to obtain a set of discrete transient measurements $\bm{H}_R$ from all paths arriving at each sensing point $\xs$, such that $\bm{H}_R$ is comparable to the captured matrix $\bm{H}$. 
To this end, we first discretize $|{\mathcal{K}}(\sPath, \sTime)|$ into neighboring bins $\tau$ using a differentiable Gaussian distribution function as
${\mathcal{\hat K}}(\sPath, \tau) =  |{\mathcal{K}}(\sPath, \sTime)|\exp\left({-\frac{(\tau-t)^2}{2\sigma_t^2}}\right)$,
where $\tau$ is a transient bin index, $t$ is continuous time of $\sPath$ (Equation~\ref{eq:tof-function}), and $\sigma_{t}$ is set to $0.62$ to make the FWHM of the Gaussian distribution cover a unit time bin. 

The time-resolved measurement $\bm{H}_r(\xl, \xs, \tau)$ at temporal index $\tau$ is then approximated as the sum of the discrete path contributions ${\mathcal{\hat K}(\sPath, \tau)}$ sampled through the concentric disk mapping as
\begin{align}
\label{eq:transport_model2}
\bm{H}_r(\xl, \xs, \tau) \approx \sum\limits_{\sPath \in \mathcal{X}} {\mathcal{\hat K}}(\sPath, \tau),
\end{align}
where $\mathcal{X}$ is the set of paths $\sPath$ that start at $\xl$ and end in  $\xs$.
\NEW{After generating the rendered transient data $\bm{H}_r$, we then apply our joint laser-sensor model to it to obtain a sensed transient data~$\bm{H}_R$:
\begin{align}
	\label{eq:transport_model3}
	\bm{H}_R(\xl, \xs, \tau) = \Psi(\tau) * \bm{H}_r(\xl, \xs, \tau) + \eta_s
\end{align}
where $\eta_s$ is the intensity offset parameter that takes the ambient light and the dark count rate of the sensor into account.}

\subsection{Optimization of system parameters}
\label{sec:optimization}
Our final goal is to estimate the system parameters~$\Theta = \{ \Theta_\mathrm{pf}, \Theta_\mathrm{ls}, \Theta_G\}$ that minimize the loss between the measured matrix $\bm{H}$ and the rendered matrix $\bm{H}_R$ (Figure~\ref{fig:full_pipeline}, red). 
We define this as
\begin{align}
\min_{\Theta} \mathcal{L}(\bm{H}, \bm{H}_R),
\label{eq:opt_measurements}
\end{align}
which we minimize by gradient descent. The transient cost function~$\mathcal{L}$ consists of a data term and regularization terms as
\begin{align}\label{eq:cost}
\mathcal{L}(\bm{H}, \bm{H}_R) =  {E_H} + {E_{{I_\mathrm{pf}}}}  + E_{\rho}.
\end{align}
\NEW{The data term $E_H$ computes an~$l_2$ norm between the transient measurements~$\bm{H}$ and~$\bm{H}_R$:}
\begin{align}
\label{eq:transient-cost}
{E_H} = \frac{1}{N_{H}}\sum\limits_i {\left\| {{\bm{H}_{i}} - \bm{H}_{R,i}} \right\|_2^2} ,
\end{align}
where $N_H$ is the total number of elements of $\bm{H}$.
The key insight of this loss term is that $\bm{H}_R$ is the byproduct of time-resolved illumination computed from our implicit geometry $G$, which was itself generated from volumetric intensities $I_\mathrm{pf}$ by means of RSD propagation of the ground truth $\bm{H}$. The difference between $\bm{H}$ and $\bm{H}_R$ is therefore a critical measure of the accuracy of geometry $G$ and $I_\mathrm{pf}$. By backpropagating the loss term through our pipeline, we optimize all system parameters, which improve the estimation of $I_\mathrm{pf}$, $G$ and therefore $\bm{H}_R$.

The term ${E_{{I}_\mathrm{pf}}}$ in Equation~\ref{eq:cost} is a volumetric intensity regularization term that imposes sparsity, pursuing a clean image:
\begin{align}
\label{eq:tv-cost}
{E_{{I_\mathrm{pf}}}} = \lambda_{1} \frac{1}{N_{{\mathrm{pf,z}}}}\sum\limits_j {\left| {I_{\mathrm{pf,z}, j}} \right|},
\end{align}
where ${I_\mathrm{pf,z}}$ is the maximum intensity values of $I_\mathrm{pf}$ projected to the $xz$ plane, $N_{\mathrm{pf,z}}$ is the number of pixels of $I_\mathrm{pf,z}$, and $\lambda_{1}$ is a loss-scale balance hyperparameter, 
which is set to 
\NEW{1e+2} in all our experiments.

The term $E_{\rho}$ in Equation~\ref{eq:cost} is a regularization term that imposes smoothness, suppressing surface reflectance noise:
\begin{align}
\label{eq:tv-albedo-cost}
	E_{\rho} = \lambda_2 \frac{1}{N_{v}}\sum\limits_m {\left | {{\nabla _{xy}}{\rho(\mathbf{x}_{v,m})}} \right |},
\end{align}
where $N_{v}$ is the number of voxels $\mathbf{x}_{v}$, and $\lambda_{2}$ is a loss-scale balance hyperparameter, 
which is set to 
\NEW{5e-3} in all our experiments. \NEW{All terms $E_H$, $E_{I_\mathrm{pf}}$, and $E_{\rho}$ of the loss function are computed over batches of the transients and voxels at every iteration.}

\section{Results}
\label{sec:results}

We implement our pipeline using PyTorch. Our code runs on an 
\NEW{AMD 7763 CPU of 2.45\,GHz equipped with a single NVIDIA GPU A100}. 3D geometry is obtained from points and normals using Poisson surface reconstruction \cite{kazhdan2013screened}. Please note that we do not perform any thresholding or masking of the data prior to this step.
\NEW{We evaluate our method on 
four real confocal datasets \textsc{Bike}, \textsc{Resolution}, \textsc{SU}, and \textsc{34}, provided by \citet{OToole:2018:confocal}, \citet{ahn2019convolutional} and \citet{Lindell2019wave}; on two real non-confocal datasets \textsc{44i} and \textsc{NLOS}, provided by \citet{Liu:2019:phasor}; and  
on four synthetic confocal datasets \textsc{Erato}, \textsc{Bunny}, \textsc{Indonesian} and \textsc{Dragon}, generated with the transient renderer by \citet{Chen2020learned}.} \NEW{The real datasets include all illumination bounces and different levels of noise depending on their exposure time. The synthetic datasets include up-to third-bounce illumination. In specific cases, we manually add Poisson noise to synthetic datasets to evaluate our robustness to signal degradation.}

\subsection{Convergence of system parameters} 
\NEW{In Figure~\ref{fig:converge-plot}, we show the convergence of our system parameters in a full optimization of the \textsc{Bike} real scene,} showing as well the final reconstruction of both volumetric intensity and geometry. 
Phasor-field kernel parameters $\Omega_\textrm{pf}$ and $\sigma_\textrm{pf}$ (first column) are responsible for improving the reconstruction quality by constructing a phasor kernel (fourth column, top) that yields high-detailed geometry. 
The laser and sensor parameters (second and third columns) improve the reconstruction of the transient measurements so that the transient simulation (fourth column, bottom, orange) resembles as much as possible the input data (blue). 
\NEW{Refer to the supplemental material for more results of the progressive optimization.}

\newcommand{\cmark}{\ding{52}}%
\newcommand{\cmk}{\color{darkgreen}{\cmark}}
\newcommand{\rcmk}{\color{red}{\cmark}}
\newcommand{\fmk}{---}

\begin{table}[t]
\centering
\caption{\label{tab:ablation-study}%
	Ablation study of the impact of each component. MSE transient loss comparison with different configurations with the \textsc{Bunny} scene \NEW{with two different albedos (Figure~\ref{fig:spatially-varying-albedo}).}}
\resizebox{0.7\linewidth}{!}{%
	\setlength{\tabcolsep}{3.6pt}%
	\begin{tabular}{cccr}
		\toprule
		\multicolumn{3}{c@{\hspace{10pt}}}{Component}  & \multicolumn{1}{c}{\multirow{2}{*}{\!MSE}} \\
		\cmidrule(lr){1-3}
		Phasor kernel & Albedo & Laser-sensor model &  transient\\
		\midrule
		\cmk  & \fmk & \fmk  & \NEW{6.817e-3} %
		\\
		\cmk  & \fmk  & \cmk   & \NEW{6.627e-3}   \\		
		\fmk  & \cmk  & \fmk   & \NEWI{2.239e-3}   \\		
		\fmk  & \cmk  & \cmk   & \NEWI{2.217e-3}   \\		
		\cmk  & \cmk  & \fmk   & \NEW{2.124e-3}   \\		
		\midrule
		\cmk  & \cmk  & \cmk  & \NEW{1.971e-3}  \\ \bottomrule
	\end{tabular}%
}%
\end{table}

We evaluate the impact of each component in our optimization pipeline: phasor kernel, albedo, and laser-sensor model, using a $256\x256\x201$ voxel volume. 
\NEWI{As Table~\ref{tab:ablation-study} shows, adding albedo and laser-sensor parameters improves the result over just using the phasor parameters, while including the three components yields the best results. The impact of optimizing albedo is the most significant in this experiment.}

\subsection{\NEW{Robustness to noise}}
\begin{figure}[t]
	\centering
	\includegraphics[width=0.8\linewidth]{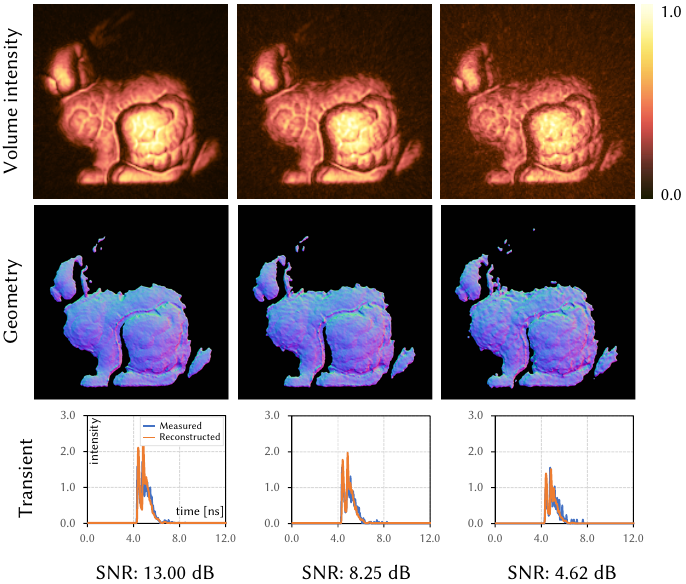}%
	\caption[]{\label{fig:result-poisson-noise_merged}
		\NEW{Evaluation of our surface reconstruction under increasing levels of Poisson noise (left to right).
			From top to bottom: intensity volume, reconstructed geometry, and measured vs. optimized transport. 
			Our method reconstructs geometry reliably across a broad spectrum of noise levels.
			A lower signal-to-noise ratio (SNR) value indicates a higher level of noise, with an exponential increase in noise.}
	}
\end{figure}
\NEW{To illustrate the robustness of our method to signal degradation, in Figure~\ref{fig:result-poisson-noise_merged} we show reconstructions of the \textsc{Bunny} synthetic dataset under increasing levels of Poisson noise (from left to right) applied to the input transient data.
The first row shows the final volumetric reconstruction after the optimization, while the second row shows the resulting surface estimation. 
The third row shows a comparison between the input transient illumination (blue) and our converged transient illumination at the same location that results from our estimated geometry (orange). 
The parameters optimized by our pipeline produce a volumetric reconstruction robust enough for our surface estimation method to obtain a reliable 3D geometry under a broad spectrum of noise levels. 
Note that while the volumetric outputs may show noticeable noise levels (first row), our pipeline optimizes the imaging parameters so that such volumetric outputs provide a good baseline for our geometry estimation method, which yields surface reconstructions that consistently preserve geometric details across varying noise levels (second row).
}

\begin{figure*}[htpb]
	\centering
	\vspace{-1mm}%
	\includegraphics[width=0.75\linewidth]{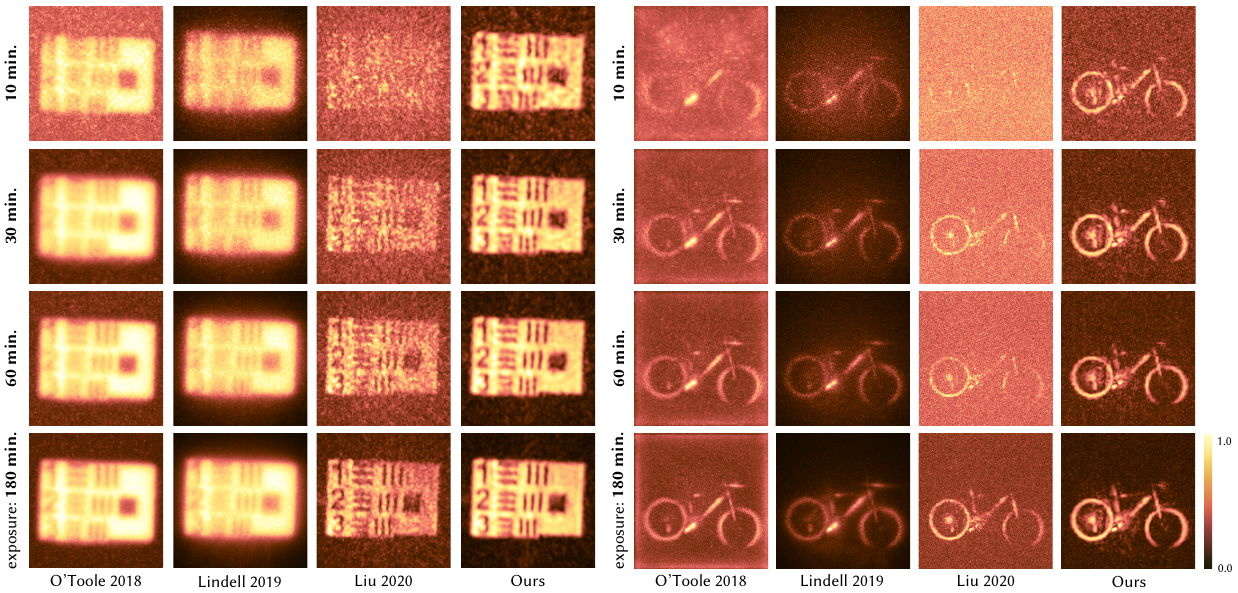}%
	\vspace{-4mm}%
	\caption[]{\label{fig:results_real_evaluation_filtering_iv}
\NEW{Reconstructed volumetric intensity comparison using the \textsc{Resolution} (left) and \textsc{Bike} (right) real scenes captured under increasing exposure times of 10, 30, 60, and 180 minutes. Existing methods \cite{OToole:2018:confocal,Lindell2019wave,Liu:2020:phasor} (first to third columns) fail to reproduce details on the resolution chart across all exposure times, and yield high-frequency noise in the reconstructions due to low SNR in the \textsc{Bike} datasets. 
Our method (last column) converges to imaging parameters that produce the sharpest results robustly under different exposure times, without requiring manual parameter tuning. 
}}
	\vspace{-3mm}
\end{figure*}

\NEW{In Figure~\ref{fig:results_real_evaluation_filtering_iv}, we compare our method with existing volumetric approaches on two real confocal scenes, \textsc{Resolution} and \textsc{Bike}, captured under different exposure times. 
	For each scene, first to fourth columns illustrate the compared methods: \citet{OToole:2018:confocal}, \citet{Lindell2019wave}, \citet{Liu:2020:phasor}, and ours, respectively. 
	First to fourth rows show the resulting volumetric intensity images under increasing exposure times of 10, 30, 60, and 180 minutes, respectively.
	Our method converges to imaging parameters that produce the sharpest results while significantly removing noise even under the lowest exposure time (top row). Other methods degrade notably at lower exposure times, failing to reproduce details in the resolution chart, or yielding noisy outputs in the \textsc{Bike} scene.  

	\NEW{While LCT \cite{OToole:2018:confocal} allows to manually select an SNR filtering parameter~$\alpha$ to improve results in low-SNR conditions, our experiments with different $\alpha$ values from $0.001$ to $1.0$ at different exposure levels 
	validate that our automated calibration approach outperforms the LCT method,  
 reproducing detailed geometric features (see supplemental material).}
	
}

\subsection{\NEW{Inverse rendering}}
\begin{figure*}[htpb]
	\centering
	\includegraphics[width=0.55\linewidth]{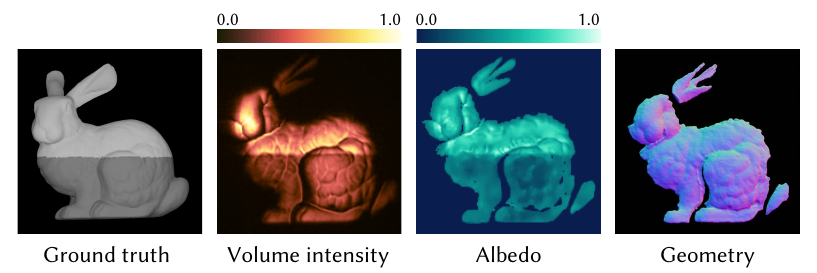}\\%
	\vspace{-4mm}
	\caption{\label{fig:spatially-varying-albedo}%
		\NEW{Our optimization scheme estimates spatially-varying albedo in a consistent manner, without affecting the surface and normal estimation. From left to right: Synthetic \textsc{Bunny} scene with two different albedos (0.3 and 1.0), our converged volumetric intensity, the optimized albedo, and the estimated geometry.  
	}}
\end{figure*}

\NEW{Our optimization pipeline estimates surface points, normals, and albedo by using only the input transient measurements.}
\NEW{Figure~\ref{fig:spatially-varying-albedo} illustrates our volumetric intensity, as well as surface points, normals and albedo in the confocal synthetic scene \textsc{Bunny} made of two different surface albedos 1.0 (top) and 0.3 (bottom). Our method is consistent when estimating spatially-varying albedo, while not affecting the estimation of detailed surface points and normals.}

\begin{figure*}[htpb]
	\centering
	\vspace{-3mm}	
	\includegraphics[width=0.6\linewidth]{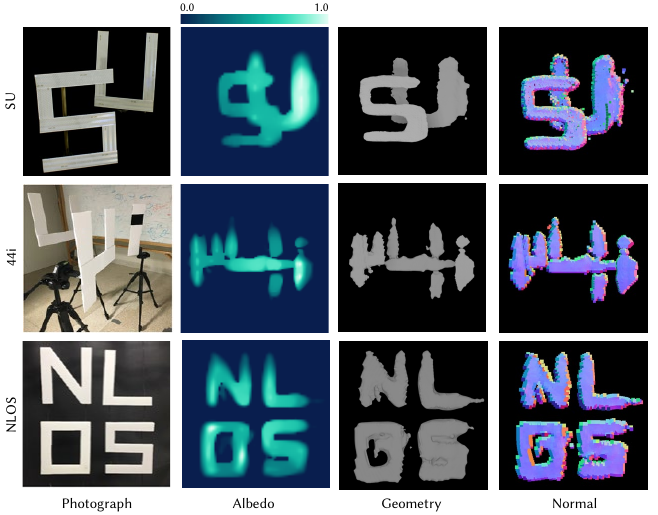}%
	\vspace{-4mm}%
	\caption[]{\label{fig:results_real_ours_geometry}
		\NEW{Our inverse rendering results with confocal real scene \textsc{SU} and non-confocal scenes \textsc{44i} and \textsc{NLOS}.
			Our approach uses transient measurements to reconstruct surface albedo (second column), geometry (third column), and normals (fourth column), estimating them correctly on multiple isolated objects at different distances. 
		}
		}
		\vspace{-5mm}
	\end{figure*}

\NEW{Figure~\ref{fig:results_real_ours_geometry} demonstrates our inverse rendering results on real scenes.
	As shown in a confocal scene \textsc{SU} (first row) and two non-confocal scenes \textsc{44i} (second row) and \textsc{NLOS} (third row), we correctly estimate the albedo of objects with uniform reflectance properties (second column), although they undergo different attenuation factors due to being at different distances from the relay wall. 
	The result of the \textsc{NLOS} non-confocal scene (third row) shows the albedo throughout the entire surface is almost identical.
	Our estimation of surface points and normals (third and fourth columns) is able to accurately reproduce the structure of the hidden geometry. 

In Figure \ref{fig:teaser}, we illustrate the benefits of our inverse rendering optimization on the real scene \textsc{Bike}. The first row shows the first iteration of the optimization, which uses the volumetric output by \citet{Liu:2020:phasor} with the default parameters of the illumination function. The resulting noise heavily degrades the geometry and normal estimation (top-right), and the albedo is wrongly estimated at empty locations in the scene despite the lack of a surface at such locations (top center). After our optimization converges (bottom row), the albedo is estimated only at surface locations, yielding a clean reconstruction of the bike's surface points and normals.}

\subsection{Geometry accuracy}
\begin{figure*}[htpb]
	\centering
	\includegraphics[width=0.9\linewidth]{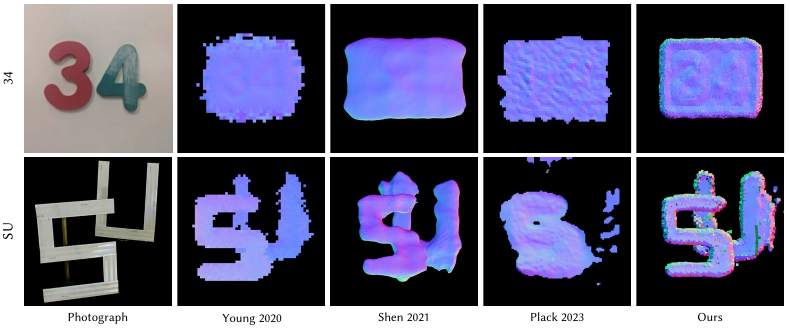}%
	\vspace{-4mm}%
	\caption[]{\label{fig:results_real_comparison_geometry}
\NEW{Comparison of the geometry estimation in two real scenes \textsc{34} and \textsc{SU}. From left to right: D-LCT~\cite{Young2020Directional}, NeTF~\cite{shen2021non}, the differential renderer \cite{plack2023fast}, and our result. Our method can reconstruct more accurately detailed features such as the depth changes of the numbers, or reproduce narrow segments of the letters, while other methods yield coarse reconstructions or even fail to reproduce partially occluded objects.}}
	\vspace{3mm}
\end{figure*}

In Figure \ref{fig:results_real_comparison_geometry}, we compare the reconstructed geometry with surface normals in two real scenes (\textsc{34} and \textsc{SU}) using D-LCT \cite{Young2020Directional}, NeTF \cite{shen2021non}, a differentiable rendering approach \cite{plack2023fast}, and our method. 
Existing methods fail to reproduce detailed surface features in both scenes, such as the subtle changes in depth of the numbers. 
\NEWI{Plack's method (fourth column) fails to reproduce the partially occluded U-shaped object and some regions of the S-shaped object in the \textsc{SU} scene. D-LCT (second column) succeeds in reproducing the U-shaped object but fails to reconstruct the detailed geometry of the boundary of the letters.} 
While NeTF \cite{shen2021non} (third column) is capable of reproducing the U-shaped object, their methodology, based on positional encoding and neural rendering, suppresses geometric details significantly, producing a coarse geometry.  
Plack's method faces similar challenges in reproducing geometric details due to the constraints imposed by the resolution of the explicit proxy geometry. 
Previous optimization-based methods that also rely on explicit geometry \cite{iseringhausen2020non, tsai2019beyond} share similar limitations.
Our method based on implicit surface representations is able to handle partial occlusions while reproducing detailed features of the surfaces, such as the depth changes on the numbers and the narrow segments of the letters.

\begin{figure*}[htpb]
\centering
\includegraphics[width=0.95\linewidth]{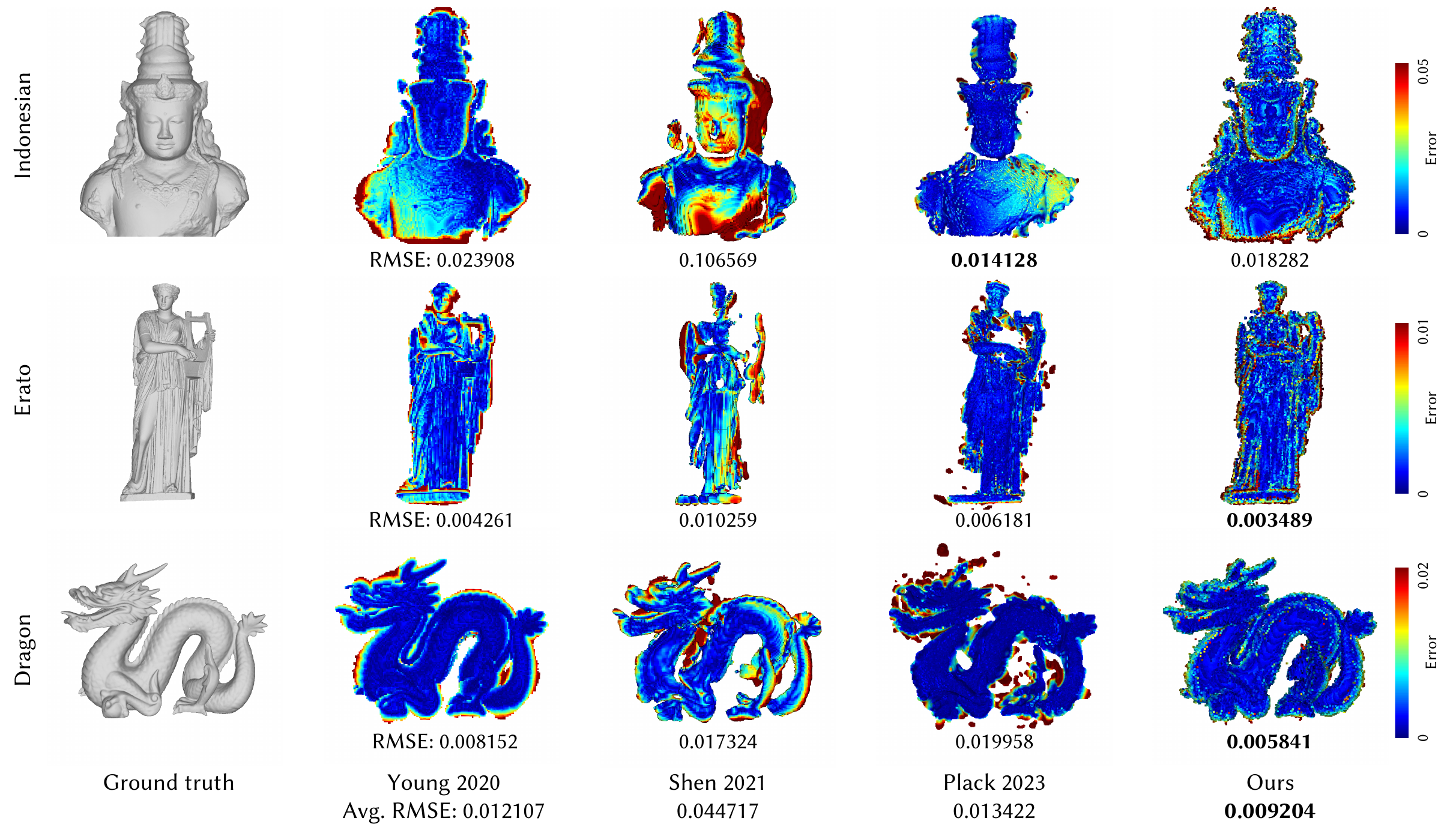}%
\vspace{-2.5mm}%
\caption[]{\label{fig:result_synthetic-geometry-comparison}
	\NEW{We perform a quantitative comparison of our surface reconstruction with \citet{Young2020Directional}, \citet{shen2021non} and \citet{plack2023fast} using synthetic transient data with ground truth geometries \textsc{Dragon}, \textsc{Erato}, and \textsc{Indonesian}.
	We quantify the introduced errors using the Hausdorff distance between the ground truth geometry and the estimated geometries.  
	Our method yields the smallest RMSE in \textsc{Erato} and \textsc{Dragon}, noticeable in highly-detailed areas. Note that while \citet{plack2023fast} has smaller RMSE in \textsc{Indonesian}, the reconstructed surface is missing significant regions of the ground truth geometry, which are not quantified by the RMSE. 
	}}
\end{figure*}

\NEW{In Figure~\ref{fig:result_synthetic-geometry-comparison}, we provide quantitative comparisons between our estimated geometry and the geometry obtained from D-LCT \cite{Young2020Directional}, NeTF \cite{shen2021non} and \citet{plack2023fast} for three synthetic scenes, \textsc{Dragon}, \textsc{Erato}, and \textsc{Indonesian}, using the Hausdorff distance map as an objective metric.
In terms of geometric accuracy, we outperform all three methods in \textsc{Erato}, and \textsc{Dragon}, as shown in the RMSE table.
Our improvements are especially noticeable in self-occluded regions and in the reproduction of detailed features. 
While \citet{plack2023fast} yields a lower RMSE in the \textsc{Indonesian} scene, note that it fails to reproduce large regions on the sides of the geometry. Thus, RMSE is only computed on the reconstructed regions and may not fully represent the overall accuracy of the reconstruction.}

\section{Discussion and Future Work}
\label{sec:discussion}
We have presented an efficient and fully-differentiable end-to-end NLOS inverse rendering pipeline, which self-calibrates the imaging parameters using only the input-measured transient illumination. 
Our method is robust in the presence of noise while 
\NEWI{achieving enhanced scene reconstruction accuracy}. %

\NEWI{Even though forward automatic differentiation (AD) is known to be memory efficient, we implemented our pipeline using reverse AD, as we found it to be 20 times faster and showed better performance when optimizing a large number of parameters (such as per-voxel albedo), and supports a wider set of differentiable functions required for our context.}

\NEWI{Phasor-field NLOS imaging can be performed analogously using temporal- or frequency-domain operators \cite{Liu:2019:phasor,Liu:2020:phasor}. However, operating in the temporal domain introduces large memory constraints that are impractical on a differentiable pipeline. Our pipeline therefore operates in the frequency domain to perform NLOS imaging, which provides practical implementation of convolutions of complex-valued phasor-field kernels within GPU memory constraints. While we based volumetric NLOS imaging on phasor-based operators and kernels, an interesting avenue of future work may be optimizing alternative kernel parameterizations or implementing other differentiable NLOS imaging approaches.}

\begin{acks} 
We want to thank the anonymous reviewers for their time and insightful comments. Min H.~Kim acknowledges the main support of the Samsung Research Funding Center (SRFC-IT2001-04), in addition to the additional support of the MSIT/IITP of Korea (RS-2022-00155620, 2022-0-00058, and 2017-0-00072), Samsung Electronics, and the NIRCH of Korea (2021A02P02-001). This work was also partially funded by the Gobierno de Aragón (Departamento de Ciencia, Universidad y Sociedad del Conocimiento) through project BLINDSIGHT (ref. LMP30\_21), and by MCIN/AEI/10.13039/501100011033 through Project PID2019-105004GB-I00. 
\end{acks}



\clearpage 
\pagestyle{empty} 


\section*{Supplementary material}

# Supplemental Document: Self-Calibrating, Fully Differentiable NLOS Inverse Rendering

Kiseok Choi
KAIST
South Korea
kschoi@vclab.kaist.ac.kr

Inchul Kim
KAIST
South Korea
ickim@vclab.kaist.ac.kr

Dongyoung Choi
KAIST
South Korea
dychoi@vclab.kaist.ac.kr

Julio Marco
Universidad de Zaragoza - I3A
Spain
juliom@unizar.es

Diego Gutierrez
Universidad de Zaragoza - I3A
Spain
diegog@unizar.es

Min H. Kim
KAIST
South Korea
minhkim@vclab.kaist.ac.kr

This supplemental document provides additional information and results in support of the primary document. Refer to Table 1 for the notations and symbols used in this paper.

## 1 JOINT LASER-SENSOR CORRELATION MODEL

The joint laser-sensor model is derived as Equation 1 following the previous related work [Chen et al. 2020; Hernandez et al. 2017].

$$\begin{aligned}
H_R &= P_{PDE} \cdot (\Phi * (\Lambda * H_r + L_a)) + L_{DCR} \\
&= (\Phi * \Lambda * H_r + \Phi * L_a) + L_{DCR} \\
&= ((E_s * G_s) * G_l * H_r + (E_s * G_s) * L_a) + L_{DCR} \\
&= ((E_s * (G_s * G_l)) * H_r + (E_s * G_s) * L_a) + L_{DCR} \quad , \quad (1) \\
&= ((E_s * G_{ls} * H_r + (E_s * G_s) * L_a) + L_{DCR} \\
&= (\Psi(t; I_l, \kappa_s, \sigma_{ls}) * H_r + L_a) + L_{DCR} \\
&= \Psi(t; I_l, \kappa_s, \sigma_{ls}) * H_r + \eta_s
\end{aligned}$$

where $P_{PDE}$ denotes the photon detection efficiency. $L_a$ is the ambient light and $L_{DCR}$ is the dark count rate. $\Phi$ is the sensor model function that can be expressed in the form of convolution between exponential function $E_s$ and Gaussian function $G_s$. $\Lambda$ is the laser function that has the shape of Gaussian $G_l$. Note that the convolution of two Gaussians $G_s$ and $G_l$ can be merged to a single Gaussian $G_{ls}$. The convolution of $E_s$ and $G_{ls}$ is then expressed as $\Psi$ that has three parameters $I_l$, $\kappa_s$, and $\sigma_{ls}$. $L_a$ and $L_{DCR}$ can be summed to a single offset value $\eta_s$. Our joint laser-sensor correlation model finally has four parameters and these values are optimized in our self-calibrating pipeline.

## 2 EXPERIMENTAL DETAILS

Table 2 summarizes the type of data (confocal or non-confocal), as well as the dimensions of the transient data, the dimensions of the reconstructed volume, the total reconstruction time, and the number of iterations before convergence; note that most of our scenes are significantly larger than previously reported results by transient optimization methods.

## 3 ADDITIONAL RESULTS

This section provides additional validations and results.

*Manual parameter adjustment vs. our self-calibration.* Figure 1 compares the estimated volumetric intensities of Bike and Resolution scenes by two different methods: the light cone transform (LCT) [O'Toole et al. 2018] and ours. To handle noise in the input dataset, we manually tweak the SNR parameter in the LCT method with a very wide range from 0.001 to 1.0. Our method yields clearer results than any of the results under the explored values for the SNR parameter of LCT, throughout all exposure levels.

*Progressive optimization results.* Figure 2 show detailed progress of the optimization in the Dragon and Erato scenes, displaying the evolution of the phasor-field kernel until the converged state. While the full optimization takes 100 iterations (1.28 hours), after only 50 iterations (39 minutes) the converged phasor-field kernel parameters already yield volumetric and geometric reconstructions very close to the converged result, while the remaining iterations refine more local details.

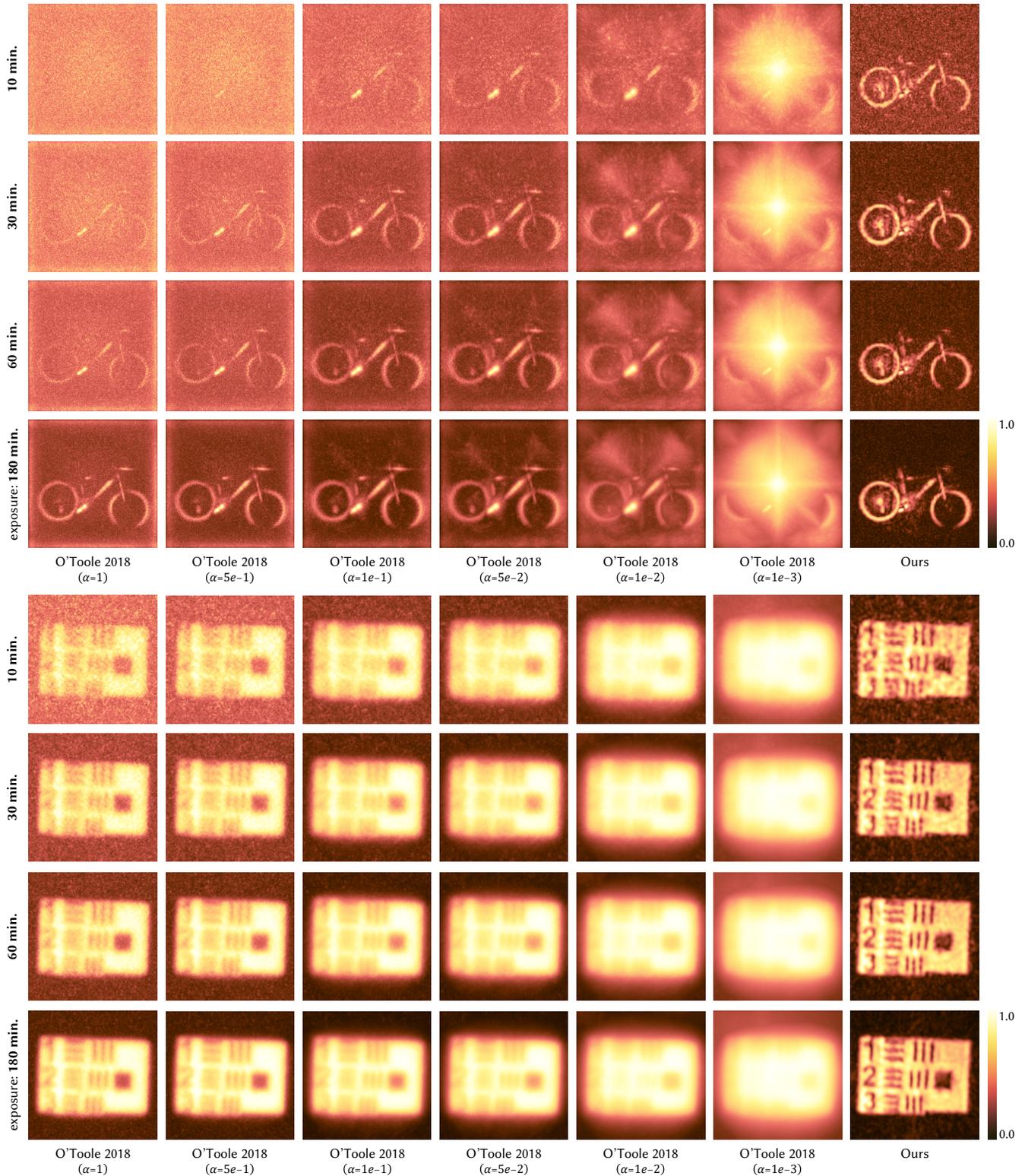

Figure 1: Comparisons of the estimated volumetric intensities of Bike and Resolution scenes by the light cone transform [O'Toole et al. 2018] and ours. To handle noise, we changed the SNR parameter $\alpha$ between 1 and 0.001 for different exposure times. In all exposure levels, our volume intensities outperform those of the LCT with manually selected parameters.

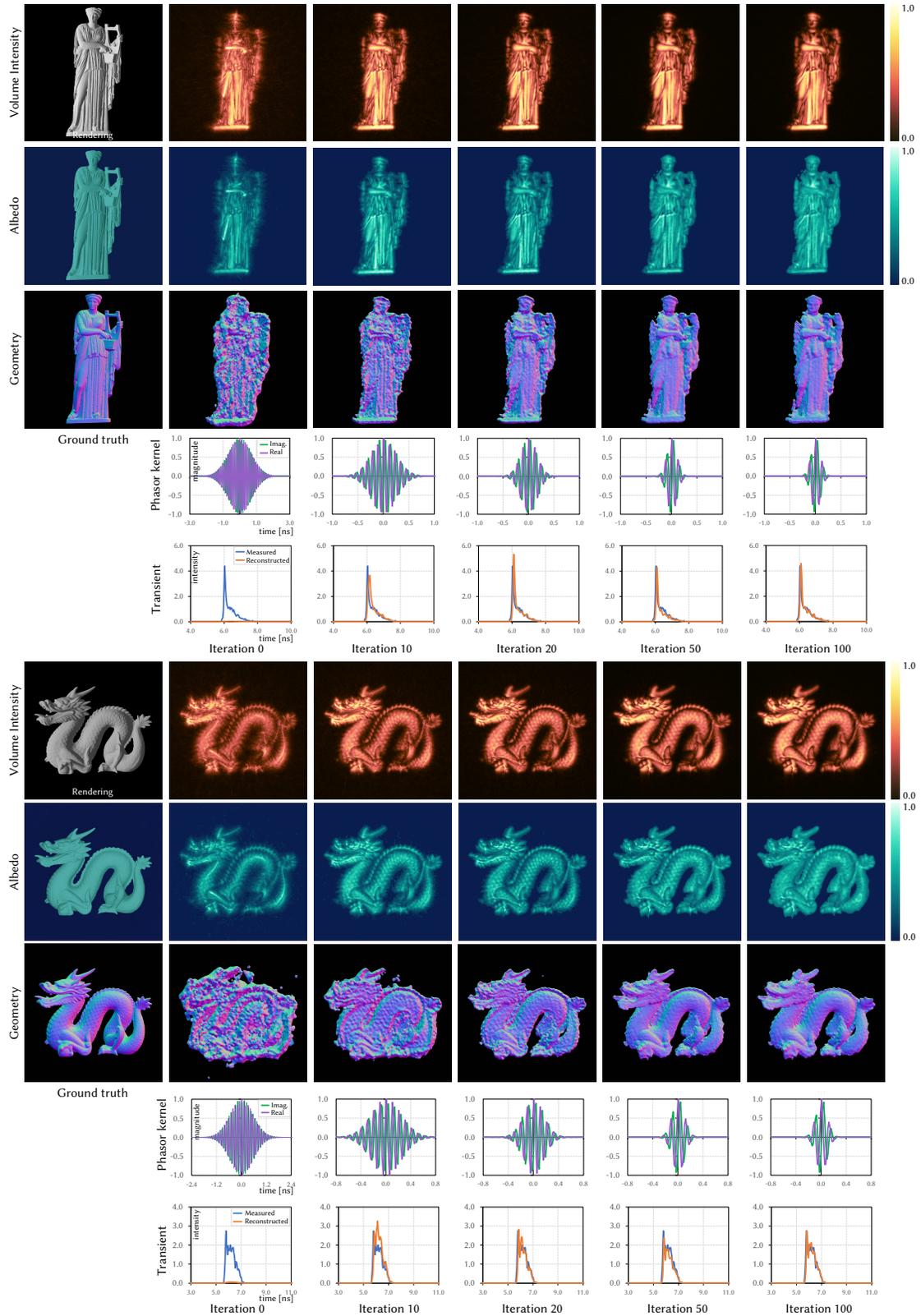

Figure 2: Progressive optimization of volumetric intensity, geometry, phasor kernel, and transient measurement samples of the Erato and Dragon scenes, showing how our reconstructions quickly converge after only 100 iterations.

Table 1: Main notations and symbols used in the paper.

| Symbol | Description |
|---|---|
| $\bar{\mathbf{x}} = \mathbf{x}_0...\mathbf{x}_k$ | Light path of $k+1$ vertices |
| $\mathbf{x}_l$ | Light source point on the relay wall |
| $\mathbf{x}_g$ | Surface point in the hidden scene |
| $\mathbf{x}_s$ | Sensor point on the relay wall |
| $\mathbf{x}_v$ | Voxel in a volumetric grid |
| $\mathbf{n}_g$ | Surface normal in the hidden scene |
| $G$ | Scene geometry parameters: points $\mathbf{x}_g$ and normals $\mathbf{n}_g$ |
| $\mathbf{t} = t_0...t_k$ | Time delays on $k+1$ vertices |
| $d$ | Distance between the hidden surface and the relay wall |
| $\psi$ | Space of all light paths |
| $\psi_k$ | Space of light paths of $k+1$ vertices |
| $\mathcal{T}$ | Space of temporal delays |
| $c$ | Speed of light in vacuum |
| tof $(\bar{\mathbf{x}})$ | Total time of path $\bar{\mathbf{x}}$ |
| $\mathcal{K}$ | Time-resolved path contribution |
| $H$ | Transient measurements |
| $H_{pf}$ | Transient measurements filtered by a phasor kernel |
| $H_r$ | Rendered transient illumination |
| $H_R$ | Rendered transient after laser-sensor model applied |
| $D\,()$ | Geometry estimation function |
| $\rho\,()$ | Reflectance function at vertex |
| $V\,()$ | Visibility function |
| $\mathfrak{T}\,()$ | Path throughput with geometric attenuation/visibility |
| $R\,()$ | Transient rendering function |
| $I_{\text{pf}}$ | Volumetric intensity backprojected by Rayleigh-Sommerfeld integrals of phasor-field diffraction |
| $\Omega_{\text{pf}}$ | Illumination frequency of phasor field kernel |
| $\sigma_{\text{pf}}$ | Illumination standard deviation of phasor field kernel |
| $\mathcal{P}\,()$ | Filtering function with a phasor field kernel |
| $I_l$ | Laser energy intensity |
| $\sigma_l$ | Standard deviation of Gaussian laser pulse signal |
| $\kappa_s$ | Sensor sensitivity decay rate |
| $\eta_s$ | Sum of ambient light and sensor dark count rate |
| $\sigma_{ls}$ | Standard deviation of Gaussian parameter for $\Psi\,()$ |
| $\Lambda\,()$ | Light source emission function |
| $\Phi\,()$ | Sensor sensitivity function |
| $\Psi\,()$ | Joint light-sensor correlation function |
| $\Theta_{\text{pf}}$ | Parameters of phasor field kernel: $\Omega_{\text{pf}}, \sigma_{\text{pf}}$ |
| $\Theta_{\text{ls}}$ | Parameters of laser and sensor models: $\sigma_{ls}, I_l, \kappa_s, \eta_s$ |
| $\Theta_G$ | Parameters of per-voxel albedo $\rho$ |
| $\Theta$ | Set of optimizing variables: $\Theta = \{\Theta_{\text{pf}}, \Theta_{\text{ls}}, \Theta_G\}$ |
| $\mathcal{L}$ | Loss function |
| $\lambda_{1...2}$ | Loss-scale balance hyperparameters |
| $\Gamma$ | Set of regularization terms |

Table 2: Configurations of our input datasets, including converge time and the number of iterations needed.

| | Scene | Confocal | Trans. measurement | Volume dimension | Time [hr. (#iter.)] |
|---|---|---|---|---|---|
| Synthetic | Bunny | Y | $256 \times 256 \times 1024$ | $256 \times 256 \times 201$ | 1.93 (100) |
| | Dragon | Y | $256 \times 256 \times 1024$ | $256 \times 256 \times 128$ | 1.28 (100) |
| | Erato | Y | $256 \times 256 \times 1024$ | $256 \times 256 \times 128$ | 1.28 (100) |
| | Indonesian | Y | $256 \times 256 \times 1024$ | $256 \times 256 \times 128$ | 1.93 (150) |
| Real | 34 | Y | $64 \times 64 \times 500$ | $64 \times 64 \times 105$ | 1.05 (300) |
| | Bike | Y | $256 \times 256 \times 512$ | $256 \times 256 \times 64$ | 1.73 (170) |
| | Resolution | Y | $256 \times 256 \times 512$ | $256 \times 256 \times 26$ | 1.33 (300) |
| | SU | Y | $64 \times 64 \times 2048$ | $64 \times 64 \times 584$ | 6.10 (200) |
| | 44i | N | $130 \times 180 \times 4096$ | $180 \times 180 \times 417$ | 3.76 (150) |
| | NLOS | N | $130 \times 180 \times 4096$ | $180 \times 180 \times 417$ | 4.38 (180) |


\clearpage 
\pagestyle{plain} 

\end{document}